# Embedding Provenance in Computer Vision Datasets with JSON-LD

Lynn Vonderhaar
*Department of Electrical Engineering and Computer Science*
*Embry-Riddle Aeronautical University*
Daytona Beach, USA
vonderhl@my.erau.edu

Timothy Elvira
*Department of Electrical Engineering and Computer Science*
*Embry-Riddle Aeronautical University*
Daytona Beach, USA
elvirat@my.erau.edu

Tyler Thomas Procko
*Department of Electrical Engineering and Computer Science*
*Embry-Riddle Aeronautical University*
Daytona Beach, USA
prockot@my.erau.edu

Omar Ochoa
*Department of Electrical Engineering and Computer Science*
*Embry-Riddle Aeronautical University*
Daytona Beach, USA
ochoao@erau.edu

*Abstract*—**With the ubiquity of computer vision in industry, the importance of image provenance is becoming more apparent. Provenance provides information about the origin and derivation of some resource, e.g., an image dataset, enabling users to trace data changes to better understand the expected behaviors of downstream models trained on such data. Provenance may also help with data maintenance by ensuring compliance, supporting audits and improving reusability. Typically, if provided, provenance is stored separately, e.g., within a text file, leading to a loss of descriptive information for key details like image capture settings, data preprocessing steps, and model architecture or iteration. Images often lack the information detailing the parameters of their creation or compilation. This paper proposes a novel schema designed to structure image provenance in a manageable and coherent format. The approach utilizes JavaScript Object Notation for Linked Data (JSON-LD), embedding this provenance directly within the image file. This offers two significant benefits: (1) it aligns image descriptions with a robust schema inspired by and linked to established standards, and (2) it ensures that provenance remains intrinsically tied to images, preventing loss of information and enhancing system qualities, e.g., maintainability and adaptability. This approach emphasizes maintaining the direct connection between vision resources and their provenance.**

## I. Introduction

As industry continues to incorporate computer vision into its processes, the importance of data provenance increases. This is especially true in safety- or reliability-critical applications where trust in the behavior of a model is crucial. Provenance allows developers to trace changes through the lifecycle of the dataset, which can help them understand the behavior of downstream models trained on that data. Data provenance also makes dataset maintenance and adaptation easier by highlighting what the dataset already contains and making it easier to understand the data gaps or modify the model's behavior. Unfortunately, provenance for image datasets is usually unavailable [1]. When it is available, it is often in separate files and rarely connects individual images to the characteristics of the full dataset. Maintaining data provenance in separate files from the data itself increases the chance of data loss over time. Conversely, embedding the data provenance within the data instances prevents data loss over time while also enhancing system maintainability.

This work proposes a novel method to connect images and dataset behaviors by embedding the provenance into the image file using JavaScript Object Notation for Linked Data (JSON-LD). JSON provides a well-known format for storing and parsing data and JSON-LD allows developers to import their existing provenance schema if they have one. This method is used to prevent data loss and enhance system maintainability and adaptability. This approach can stand alone or complement other data organization and description methods, e.g., data cards or graphs, to improve the effectiveness of data provenance.

The contributions of this paper are as follows:
1. Embedding computer vision data provenance into the data itself using JSON-LD.
2. Connecting model behavior to individual instances in the training data.
3. Enhancing model maintainability and adaptability via embedded provenance.

## II. Background

The proposed approach involves using a Linked Data construct to embed provenance into vision resources. The present section provides the background information necessary to understand the approach.

Fig. 1. Example data card format [4].

## A. Provenance

The World Wide Web Consortium (W3C) defines provenance as: "…information about entities, activities, and people involved in producing a piece of data or thing, which can be used to form assessments about its quality, reliability or trustworthiness" [2]. Provenance is the pedigree of some resource, establishing its history, meaning, and intent. With the ubiquity of computer vision in industry, especially with tasks continuously evolving in complexity, AI model transparency and explainability are concerns of increasing importance [3]. Since an AI model, e.g., some ML image recognition model, effectively embodies, or operationalizes, its training dataset(s), model transparency may be addressed via data transparency. Provenance can increase model transparency by providing a timeline of data modifications and adjustments from origin to current state [3, 1]. Data modifications include data addition, preprocessing, synthesis, annotation, etc. Provenance may even include business interests or downstream concerns, e.g., dataset requirements that the image meets [4]. Tracking and understanding these aspects of the data may help developers using the data understand the behavior of their downstream models.

Provenance has been introduced to image datasets previously in the form of data cards [4, 5]. Several authors have proposed various formats. One potential format is shown in Fig. 1 [4]. Data cards, like model cards, are a form of documentation for ML developers. Dataset documentation provides transparency and clear understanding of the purpose and capabilities of a dataset [5]. As is shown in Fig. 1, a data card may include labels such as intended uses, related documents, and creation requirements, among others. Data cards provide an overview of the dataset as a whole but do not extend to describing the origin and modifications of individual data instances.

The cultivation of data provenance has many benefits, including:

- Experiment reusability and reproducibility [6]
- Data trust, authenticity and auditability [7]
- Performance prediction [8]
- Process optimization [9]
- Fault tolerance [10]

TABLE I. METADATA AND PROVENANCE COMPARED.

| Aspect | Metadata | Provenance |
|---|---|---|
| Definition | Descriptive information about data | Origin and changes of data |
| Focus | Data characteristics: format, structure, size, etc. | Processes, actions, transformations |
| Purpose | Organization, discovery | Traceability, accountability, trust |
| Examples | Author, creation date, file size | Who modified data and how |
| Use Cases | Organizing, searching, interpreting | Ensuring authenticity, tracing errors, workflows |

Provenance is not to be confused with metadata. In short: metadata is descriptive information about some resource, while provenance is process-oriented metadata delineating how data was created and changed over time. An example of image metadata for computer vision tasks could be what objects are in some image, and how many of each. A description of the difference between provenance and metadata is shown in Table I.

*B. Linked Data*

In 1999, the United States Defense Advanced Research Projects Agency (DARPA) started the DARPA Agent Markup Language (DAML) program, which sought to provide machine-readable data for the Web [11]. DAML also intended to fulfill the Semantic Web vision of linking concepts into a universal web [12]. There are several foundational constructs facilitating the Semantic Web vision, e.g., the Resource Description Framework (RDF), RDF Schema (RDFS), the Web Ontology Language (OWL), and more [13].

Linked Data is a specific practice that is part of the Semantic Web vision, embodying a set of principles and best practices for publishing and connecting data on the Web, engendering accessibility, discoverability, reuse, etc. There are four defined principles of Linked Data [14]:

1. Use URIs as names for things
2. Use HTTP URIs so that people can look up those names
3. When someone looks up a URI, provide useful information, using the standards (RDF*, SPARQL)
4. Include links to other URIs so that they can discover more things

The approach presented in the present paper aims to include these four Linked Data principles in vision tasks by using JSON-LD and a URI-based schema to represent image provenance.

III. APPROACH

This paper uses JSON-LD to embed provenance directly into image files. This maintains the provenance of individual images, as opposed to other data provenance methods, e.g., data cards, which only track the dataset as a whole. Maintaining individual provenance decreases the chance of data loss and improves the maintainability and adaptability of the dataset by increasing the transparency of behavior origins.

The proposed approach tracks data provenance from requirements through collection, transformations, and revisions, and stores all of the data within the image itself. This schema can be used with both real and synthetic images. The instance-level view allows ML developers to understand how each image fits into the dataset and contributes to a downstream model's behavior. The provenance embeddings could even help in sorting large company datasets based on behaviors that a downstream model requires.

*A. Types of Provenance Data*

As was mentioned previously, provenance is the description of data origins and lifecycle. Therefore, it includes origin information, e.g., method of collection and date of collection, but also transformation information, e.g., resizing, normalizing, and annotating. A full description of data provenance is shown in Table II.

As is shown in Table II, there are many different types of data provenance to save. The choice of what data to save for each application is up to the developer. It is possible that in some cases, all collected data was used, in which case inclusion and exclusion criteria may not be relevant.

The more data that is saved from the development process, the better the transparency of model behavior is. It is important to note that data origins may include dataset requirements [15]. Because ML is data-driven, as developers note what behaviors their model should exhibit, it defines what behaviors the data should show. Saving this information within the images themselves can help developers sort data based on what behaviors they impart to downstream models [15].

TABLE II. TYPES OF DATA PROVENANCE.

| Data Origins | Data Transformations | Data Revisions |
|---|---|---|
| • Dataset requirements [15]<br>• Datasets [20]<br>• Data type<br>• Data source<br>• Data fidelity (real or synthetic)<br>• Data metadata, as applicable (resolution, size/length, contrast, etc.)<br>• Synthetic data metadata, as applicable (prompt, seed, etc.)<br>• Method of collection, collector, date of collection [20]<br>• Inclusion/exclusion criteria (person, date) | • Preprocessing steps (transformations, cleaning, denoising, resizing, cropping, normalization) (person and date for each) [20]<br>• Labeling (annotation type, annotator, date of annotating) [20]<br>• Feature selection, if relevant (person and date) [20]<br>• Training/validation/testing split<br>• Training/validation/testing dataset class proportions | • Data changes, e.g., adding data, removing data<br>• Dataset versioning [20]<br>• Reverting to previous versions [20] |

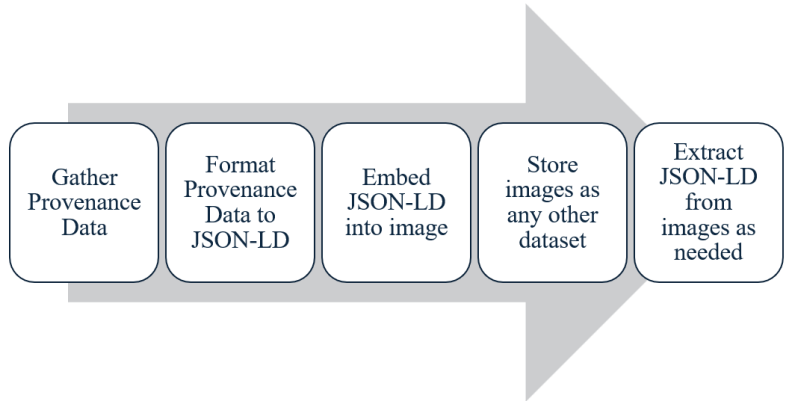

Fig. 2. The steps proposed process from gathering provenance to utilizing it.

*B. Method*

The method proposed in this work uses JSON-LD to format image provenance data before embedding it into the image and storing it within the dataset. The exact schema used to store the data provenance and the data saved may vary for different applications and developers, but the proposed process remains the same. The full process is shown in Fig. 2.

Once the important provenance data is gathered, the pipeline formats it into JSON-LD, which enables alignment with existing organization schemas and easy querying based on JSON tags. The JSON-LD description is then stored as the image metadata and the image may then be stored as an instance in the dataset, just like any other computer vision dataset. When needed, the JSON-LD data may be extracted from the image and queried or used as needed. The code for this method can be found at https://github.com/lynndalou/JSONLD.

## IV. RESULTS

The example results for this method use a schema based on Schema.org, which is a collaborative effort between Google, Microsoft, Yahoo, and Yandex to provide open source linked data schemas [16]. The schema used in this experiment is based on the ImageObject from Schema.org. The schema is then expanded using a custom schema for images generated using Flux. This custom schema can be found at https://github.com/lynndalou/JSONLD.

The provenance data was collected manually for this proof of concept. However, this collection method could easily be automated by extracting the necessary parameters from Flux, formatting them, and embedding them into the generated image. The provenance was embedded into the image as image metadata. The implementation for this method using Python and JSON-LD can be found at https://github.com/lynndalou/JSONLD.

An example of this method is shown in Figs. 3 and 4. Fig. 3 shows an image of two dogs in a dog park that was generated using Flux. Its provenance was saved as image metadata using the method described in the approach section, and was then extracted from the image, as shown in Fig. 4.

## V. DISCUSSION

The method presented in this work allows image provenance to be saved directly with the image that it describes. This has several benefits, including decreasing data loss and increasing the maintainability of the computer vision dataset. Since the provenance is saved directly within the image, as opposed to within a separate file, this method prevents data loss. Additionally, the dataset becomes more maintainable and adaptable because the images can be sorted by the behaviors that they impart to downstream ML models. If behaviors, e.g., dataset requirements in the case of the show results, are saved within the data provenance, then datasets can be sorted based on the requirements that they satisfy. This makes the dataset more maintainable, as it becomes easier to identify or address gaps. It also makes the dataset more adaptable, as the images imparting a specific behavior may be chosen for a new

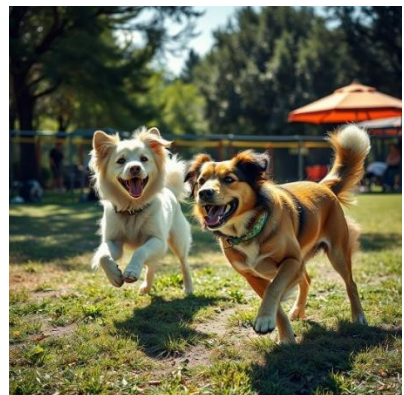

Fig. 3. Image generated using Flux.

```
Extracted text:
{
"@context": "https://schema.org",
"@type": "ImageObject",
"flux": "https://example.org/flux"
"name": "image_of_person"
"creator": {
        "@type": "Person",
        "name": "Author 1"
}
"methodOfCollection": "flux",
"dateCreated": "2025-03-02T09:31:00Z",
"encodingFormat": "image/png",
"flux:version": "flux1.schnell",
"flux:parameters": {
        "prompt": "A photo taken with a Nikon Z9 of two dogs playing in a vibrant dog park on a sunny afternoon.",
        "seed": "140716430322376",
        "steps": "4",
        "sampler": "euler_ancestral",
        "width": "1024",
        "height": "1024"
}
"requirements": {
        "requirement": "The object detector shall detect a "Dog" class when the class is in a park setting",
        "requirement": "The object detector shall detect a "Dog" class when there are 2 instances of the class in the image"
}
"annotations": {
        "class": "Dog",
        "bbox": "[x1, y1, x2, y2]",
        "class": "Dog",
        "bbox": "[x1, y1, x2, y2]"
}
}
```

Fig. 4. Extracted provenance data from Fig. 3.

application while others may be left behind. Testing this method for maintainability will be addressed in future work.

The benefits of specifically using JSON-LD over other formats is that it becomes easy to transfer existing schemas to this use case. Schemas based on ontologies and linked data can be referenced and later used to sort the data as needed. This method could, however, be adjusted as needed to fit the needs of a specific organization or developer. The proposed method works with any format of the data that suits the application's needs. Adaptations could include using standard JSON format, or another organization-defined format. Additionally, this method was shown both for real data and synthetic data, though the schema and important provenance differs slightly between the two data collection methods.

The results shown in this work are used in isolation. However, this method could be used in conjunction with other provenance saving methods, e.g., data cards to improve the maintainability of the system. For example, if using in conjunction with data cards, the provenance saved using this method would describe the provenance for each instance within the dataset while the data card would save provenance, behavior, and usage data for the dataset as a whole. It is up to the organization or developer how much provenance data they need, but using this method in conjunction with other methods has its benefits.

## VI. RELATED WORK

The reproducibility of ML workflows, even in the scientific landscape, is suspect, due to a lack of provenance. A great body of work pertaining to ML provenance exists. For one, Moreira et al. present an end-to-end processing pipeline for image provenance analysis, citing the use case of estimating the types of transformations some image has undergone from prior ones, e.g., to identify fake or edited images; their system produces a provenance graph [17].

Provenance is also becoming an increasingly prominent topic in AI development. Longpre et al. discuss AI data authenticity and consent in the age of large-scale generative AI, focusing on the need for standardized provenance [7]. Samuel et al. introduce ProvBook, a means of storing and sharing ML provenance in Jupyter Notebooks, citing the FAIR (Findable, Accessible, Interoperable, Reusable) principles for data fairness [18]. Provenance is also found to be helpful in making AI more transparent and reproducible [18, 19]. By tracking each step of the development process, including data collection and processing, model hyperparameter tuning, model training and testing, and more, tracking the model's behavioral development becomes easier. Many of these papers discuss what provenance should be saved, but none use JSON-LD to save provenance without the risk of data loss.

## VII. Limitations and Future Work

The example results shown in this paper are manually achieved. Continuing to follow this method manually would limit its scalability. However, developing a provenance collecting and saving pipeline would be trivial, e.g., downloading, formatting, and saving metadata from synthetic data generation. However, even if the method is automated, scaling the method to hundreds of thousands of images may still be difficult due to the sheer amount of data that is stored. Not only would capturing that amount of data be difficult but parsing it would take a significant amount of time and computational power. Future work will explore solutions to this problem.

Future work will also analyze the feasibility of data reuse for applications that follow a different schema. Of course, the developer or organization is not required to use a schema, but if they do, reusage of data into a different schema may be limited. If the new schema requires information that was not captured initially, data reuse may not be possible.

To help address the first two points, future work will further analyze the provenance categories and analyze potential areas for refinement or merging. This could help reduce the amount of data saved, and reduce the variation in data saved between applications, thereby addressing both of the above points.

This work focuses on embedding provenance into images, but a similar approach could be used for videos, gifs, or other digital media. Future work will discuss how the provenance categories and the proposed method can adapt to work for these files as well. Finally, future work will explore the possibility of encrypting the JSON-LD provenance data as a method of preventing data tampering. This exploration will also extend to protecting videos and other digital media.

## VIII. Conclusion

The purpose of this paper was to provide a solution to data loss and maintainability problems when saving provenance for computer vision datasets. This was accomplished by saving the image provenance as metadata within the image itself. The image provenance was formatted as JSON-LD for easier transferability from existing provenance schemas. Because the provenance data is stored within the image itself, this method prevents data loss, which is common when data is stored in a separate file. It also makes maintainability and adaptability of the dataset to other applications easier because the data can more easily be sorted based on the behavior that it imparts to downstream ML models. Future work for this method includes an exploration of the scalability of this method and a deeper analysis of data reuse with this metho. Additionally, future work will expand this method to videos, gifs, and other digital media formats.